\documentclass{article}
\usepackage{iclr2020_conference,times}

\usepackage{amsmath,amsfonts,bm}

\def\eqref#1{equation~\ref{#1}}
\def\1{\bm{1}}

\DeclareMathAlphabet{\mathsfit}{\encodingdefault}{\sfdefault}{m}{sl}
\SetMathAlphabet{\mathsfit}{bold}{\encodingdefault}{\sfdefault}{bx}{n}

\def\sR{{\mathbb{R}}}

\usepackage{hyperref}
\usepackage{url}

\usepackage[pdftex]{graphicx}
\usepackage{multicol}

\title{Revisiting Fine-tuning for Few-shot Learning}

\author{
	Akihiro Nakamura \\
	The University of Tokyo \\
	\texttt{nakamura@mi.t.u-tokyo.ac.jp} \\
	\And 
	Tatsuya Harada \\
	The University of Tokyo, RIKEN \\
	\texttt{harada@mi.t.u-tokyo.ac.jp} \\
}

\iclrfinalcopy 
\begin{document}

\maketitle

\begin{abstract}
Few-shot learning is the process of learning novel classes using only a few examples 
and it remains a challenging task in machine learning.
Many sophisticated few-shot learning algorithms have been proposed
based on the notion that networks can easily overfit to novel examples
if they are simply fine-tuned using only a few examples.
In this study, we show that in the commonly used low-resolution {\it mini}-ImageNet dataset, 
the fine-tuning method
achieves higher accuracy than common few-shot learning algorithms in the 1-shot task
and nearly the same accuracy as that of the state-of-the-art algorithm in the 5-shot task.
We then evaluate our method with more practical tasks, namely
the high-resolution single-domain and cross-domain tasks.
With both tasks, we show that our method achieves higher accuracy than common few-shot learning algorithms.
We further analyze the experimental results and show that:
1) the retraining process can be stabilized by employing a low learning rate, 
2) using adaptive gradient optimizers during fine-tuning can increase test accuracy, and
3) test accuracy can be improved by updating the entire network 
when a large domain-shift exists between base and novel classes.
\end{abstract}

\section{Introduction}
Previous studies have shown that high image classification performance can be achieved 
by using deep networks and big datasets \citep{alex_net, VGG, ResNet, google_net}.
However, the performances of these algorithms rely heavily on extensive manually annotated images,
and considerable cost is often incurred in preparing these datasets.
To avoid this problem, few-shot learning,
which is a task of learning novel classes using only a few examples, has been actively researched.

However, few-shot learning remains a considerably challenging task in machine learning,
and classification accuracy in few-shot tasks is much lower than that of the {\it many}-shot regime.
This is because a network pretrained using base classes
must adapt to novel classes using only a few examples.
The simplest means of overcoming this difficulty is to fine-tune the network using novel classes.
However, the number of trainable parameters of deep networks is so large
that we believe that networks can easily overfit to novel classes
if we simply fine-tune the networks using only a few examples.
For example, the number of trainable parameters in the ResNet-152 \citep{ResNet} is approximately 60 M,
which is much greater than the number of novel examples (e.g., 25 for 5-way 5-shot learning),
and this leads us to the idea of overfitting.
Using various sophisticated methods, numerous studies have been conducted to prevent networks from overfitting.
However, the performance of a naive fine-tuning method has not been well investigated, and
\cite{closer_look} has pointed out that performance of this method had been underestimated in previous studies.
Therefore, in this study, we analyze the performance of a fine-tuning method
and show that it can achieve higher classification accuracy than common few-shot learning methods
and, in some cases, can achieve an accuracy approximating that of the state-of-the-art algorithm.
We also experimentally show that: 
1) a low learning rate stabilizes the retraining process,
2) using an adaptive gradient optimizer when fine-tuning the network increases test accuracy, and
3) updating the entire network increases test accuracy when a large domain shift occurs between base and novel classes.

To evaluate accuracy in few-shot image classification tasks,
the {\it mini}-ImageNet dataset \citep{matching_net} has been used in many previous studies.
This is a subset of the ImageNet dataset \citep{imagenet}
in which each image is resized to $84 \times 84$ to reduce computational cost.
Recently, \cite{closer_look} evaluated few-shot learning methods in more practical datasets,
namely, the high-resolution {\it mini}-ImageNet dataset and cross-domain dataset.
Both datasets contain higher-resolution images than the original {\it mini}-ImageNet dataset,
and the cross-domain dataset represents a greater challenge 
because base and novel classes are sampled from different datasets.
Thus, a larger domain shift occurs between these classes.
In this study, we evaluate the performance of our method 
using the high-resolution {\it mini}-ImageNet dataset (high-resolution single-domain task) and
cross-domain dataset (cross-domain task) 
as well as the common low-resolution {\it mini}-ImageNet dataset (low-resolution single-domain task).
Details of these datasets are provided in Section \ref{sec:datasets}.

The main contributions of this study are as follows:

1) We show that in the common low-resolution single-domain task, 
our fine-tuning method achieves 
higher accuracy than common few-shot learning algorithms in the 1-shot task
and nearly the same accuracy as that of the state-of-the-art method in the 5-shot task.
We also show that our method achieves higher accuracy than common few-shot learning methods
both in the high-resolution single-domain and cross-domain tasks.
Note that we do not compare the performance of our method with the state-of-the-art algorithm in 
the high-resolution single-domain and cross-domain tasks
because the performances for these tasks are not reported in the corresponding papers.

2) We further analyze the experimental results and show 
that a low learning rate stabilizes the relearning process,
that test accuracy can be increased by using an adaptive gradient optimizer such as the Adam optimizer,
and that updating the entire network can increase test accuracy when a large domain shift occurs.

\section{Overview of Few-shot Learning}
\subsection{Notation}
Few-shot learning is a task of learning novel classes using only a few labeled examples.
This task is also called $N$-way $K$-shot learning,
where $N$ denotes the number of novel classes 
and $K$ is the number of labeled examples per class.
We focus on the 5-way learning task such as in previous studies \citep{closer_look, delta_encoder}.
Labeled and unlabeled examples of novel classes are called support and query sets, respectively.
A network is pretrained using base classes, which contain numerous labeled examples.
Base and novel classes are mutually exclusive.
Base classes are used for pretraining,
and novel classes are used for retraining and testing.
Validation classes are used to determine a learning rate and the number of epochs required to retrain the network.

\subsection{Few-shot Learning Algorithms}
To date, numerous few-shot learning algorithms have been proposed,
and these methods can be roughly classified into three categories:
learning discriminative embedding using metric-based classification,
learning to learn novel classes, and
data-augmentation using synthetic data.

{\bf Metric-based Approach} \\
Metric-learning approaches such as MatchingNet \citep{matching_net} and ProtoNet \citep{proto_net}
tackle few-shot classification tasks by training an embedding function 
and applying a differentiable nearest-neighbor method to the feature space using the Euclidean metric.
RelationNet \citep{relation_net} was developed to replace the nearest-neighbor method with a trainable metric 
using convolutional and fully connected (FC) layers
and has achieved higher few-shot classification accuracy.
\cite{imprinted_weight} proposed a method called weight imprinting.
They showed that including normalized feature vectors of novel classes in the final layer weight
provides effective initialization for novel classes.
We use the weight-imprinting method to initialize the last FC layer before fine-tuning the network.
These conventional methods successfully retrain networks using novel classes while preventing overfitting,
but we show that few-shot classification performance can be further improved by fine-tuning networks.

{\bf Meta-learning-based Approach} \\
Meta-learning-based approaches address the few-shot learning problem
by training networks to {\it learn to learn} novel classes.
\cite{opt_as_model} focused on the similarity between gradient descent methods and long short-term memory (LSTM) \citep{LSTM},
and they achieved a fast adaptation to novel classes by using LSTM to update network weights.
\cite{MAML} proposed a method to train a network to obtain easily adaptable parameters
so that the network can adapt to novel classes 
by means of a few gradient steps using only a few examples.
In these algorithms, networks are explicitly trained to learn how to adapt to novel classes.
However, we show that networks pretrained without explicit meta-learning methods
can also learn novel classes and achieve high few-shot classification accuracy.

{\bf Data-augmentation-based Approach} \\
Data-augmentation-based approaches overcome data deficiencies
by generating synthetic examples of novel classes.
Some methods synthesize examples of novel classes 
by applying within-class differences of base classes to real examples of novel classes \citep{hallucination, delta_encoder}.
\cite{imaginary_data} integrated a feature generator using a few-shot learning process
and succeeded in generating synthetic data using only a few novel examples.
These methods succeeded in improving the performance of few-shot learning by using synthetic examples for retraining.
Nevertheless, we show that networks can adapt to novel classes 
by using only naive data-augmentation methods such as image flipping and image jittering.

\subsection{Datasets for Few-shot Learning}
\label{sec:datasets}
The {\it mini}-ImageNet dataset is a well-known dataset used to evaluate few-shot learning methods.
The dataset was first proposed by \cite{matching_net},
but the train/validation/test split proposed by \cite{opt_as_model} is often used instead.
Therefore, we used this split in this study.
This dataset is a subset of the ImageNet dataset \citep{imagenet}
and contains 100 classes with 600 examples of each class.
The classes are split into 64 base, 16 validation, and 20 novel classes.
Images in this dataset are resized to $84\times84$ to reduce computational cost.

Recently, \cite{closer_look} used a higher-resolution {\it mini}-ImageNet dataset 
with an image resolution of $224\times224$ to employ deeper networks.
They also revealed that the domain shift that occurs between base and novel classes in the {\it mini}-ImageNet dataset is small
because the classes are sampled in the same dataset.
The authors proposed the cross-domain dataset, which has a larger domain shift between these classes.
In this dataset, the whole {\it mini}-ImageNet dataset is used as a set of base classes,
and randomly sampled 50 and 50 classes from the CUB-200-2011 dataset \citep{CUB} are used 
as validation and novel classes, respectively.
These datasets are more practical because they use high-resolution images,
and the cross-domain dataset is more challenging because the domain shift that occurs between base and novel classes is larger.
Therefore, we use the high-resolution {\it mini}-ImageNet dataset and cross-domain dataset for evaluation
as well as the common low-resolution {\it mini}-ImageNet dataset.

\section{Experiments}
\subsection{Network Structure}
\label{sec:network_structure}
\begin{figure}[t]
	\centering
	\includegraphics[width=\linewidth]{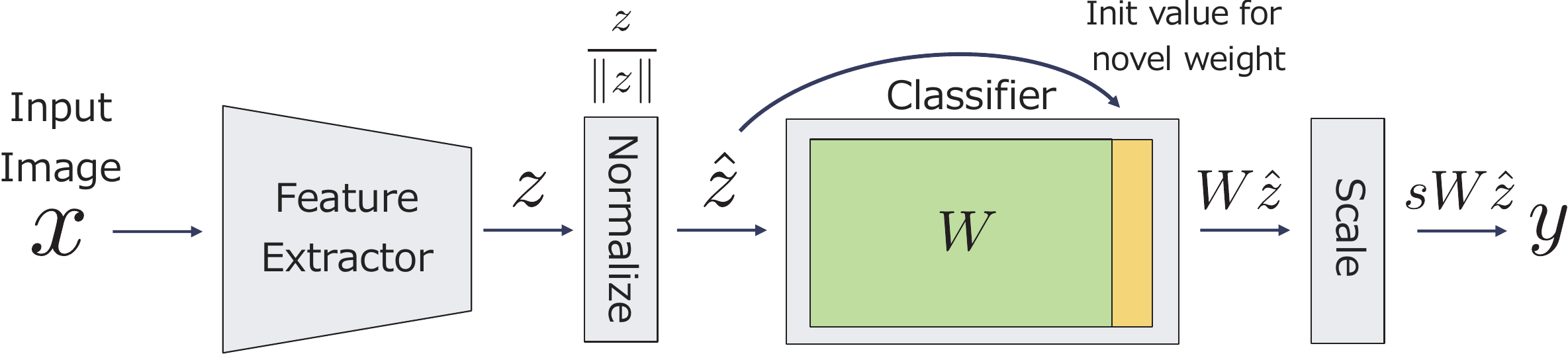}
	\caption{
		For the normalized classifier, we use the technique called weight imprinting proposed by \cite{imprinted_weight}.
		Each column $\bm{w}_i \in \sR^d$ of the classifier weight is normalized
		so that $w_i$ has a norm of 1 (i.e., $\|w_i\| = 1$),
		and classification is performed by taking the inner products between $w_i$
		and normalized feature vector $\hat{\bm{z}}\in \sR^d$.
		Note that variable $d$ is the dimension of the feature space.
		Before the network is fine-tuned, the initial weight for a novel class can be obtained
		by including feature vector $\hat{\bm{z}}$ of the novel class in the classifier weight.
		When multiple novel examples per class are available (i.e., $K$-shot learning with $K>1$),
		the initial weight for the novel class can be obtained
		by normalizing class mean $1/K \sum_{j=1}^{K} \hat{\bm{z}}_j$ again.
		Because the output range of $W\hat{\bm{z}}$ is $[-1, 1]$,
		ensuring that the probability of the correct label approximates $1$ using softmax activation is difficult.
		This problem can be avoided by applying scale factor $s\in \sR$ to the output,
		as discussed by \cite{imprinted_weight}.
	}
	\label{fig:normalized_classifier}
\end{figure}
The ResNet-18/34/50/101/152 \citep{ResNet} and VGG-16 \citep{VGG} without FC layers 
are used as feature extractors in this study.
Note that the last MaxPool2d layer of the VGG-16 is replaced by the GlobalAveragePool2d layer 
to support different resolutions of input images.
We also use the simple classifier (i.e., common FC layer) and the normalized classifier.
The technique known as weight imprinting \citep{imprinted_weight} is used in the normalized classifier;
the normalized classifier is illustrated in Figure \ref{fig:normalized_classifier}.
Before fine-tuning the normalized network for novel classes, we initialize
classifier weight $W$ by deleting the weight and
inserting columns for novel classes, as shown in Figure \ref{fig:normalized_classifier}.
Regarding the simple classifier, the initial weight for novel classes can be obtained 
by applying the multi-class linear SVM to feature vectors of novel classes.

\subsection{Dataset}
We evaluated our method using
the low-resolution {\it mini}-ImageNet dataset as a common evaluation dataset.
In addition, we used the high-resolution {\it mini}-ImageNet and cross-domain datasets
as more practical datasets.
Details of these datasets are provided in Section \ref{sec:datasets}.
In this study, we identify tasks that use these datasets as
low-resolution single-domain task,
the high-resolution single-domain task, and
cross-domain tasks.

\subsection{Pretraining the Network}
\label{sec:pretraining}
The networks were pretrained by using the base classes of the datasets for $600$ epochs.
We used the Adam optimizer \citep{Adam-Adamax} 
with a learning rate of $0.001$ in the same manner as \cite{closer_look}.
These parameters for pretraining are normally optimized using validation classes,
but we fixed these parameters to reduce computational cost.
Input images were preprocessed by
random-resized cropping with a size of $224\times224$.
We also performed color jittering and
random-horizontal flipping, and we
subtracted channel-wise means of the ImageNet dataset $(0.485, 0.456, 0.406)$.
In addition, division by channel-wise standard deviations of the ImageNet dataset $(0.229, 0.224, 0.225)$
was performed in the same manner as \cite{closer_look}.
Note that in the low-resolution single-domain task,
the size of the random-resized cropping was set to $84\times84$.

\subsection{Updating the Network for Novel Classes}
In this study, we compared three fine-tuning methods in which:
1) the entire network is updated,
2) the classifier weight and batch-normalization (BN) statistics are updated, and
3) only the classifier weight is updated.
The third method is a common fine-tuning method to prevent overfitting.
The second method is based on a previous study \citep{batch_statistics}
that successfully fine-tuned an image generator to a novel class without overfitting
by updating only the BN statistics (i.e., $\gamma$ and $\beta$ of BN layers).
A similar approach is known as meta-transfer learning (MTL) \citep{MTL}.
The authors who proposed MTL showed that updating only scales and biases of network parameters
prevents them from overfitting while achieving efficient adaptation to unseen tasks.
Although the methods proposed by \cite{batch_statistics} and \cite{MTL} presented similar ideas,
we chose the former because of its simplicity in implementation.

Initial classifier weights for novel classes were obtained before we fine-tuned the networks,
as discussed in Section \ref{sec:network_structure}.
The networks were retrained with mini-batch-based learning
with a batch size of $NK$ in the $N$-way $K$-shot learning scenario.
The learning rate and number of epochs for fine-tuning were determined by using validation classes.
We evaluated few-shot classification accuracy by calculating 
the mean accuracy of $600$ trials 
using randomly sampled classes and examples in the novel classes.
We also calculated the $95\%$ confidence interval of the mean accuracy.
In the validation, test, and network initialization phases for the novel classes,
input images were preprocessed by
resizing to $256\times 256$,
center-cropping to a size of $224\times224$,
subtracting channel-wise means of the ImageNet dataset $(0.485, 0.456, 0.406)$,
and dividing by channel-wise standard-deviations of the ImageNet dataset $(0.229, 0.224, 0.225)$
in the same manner as \cite{closer_look}.
The input preprocessing for fine-tuning phase was the same as discussed in Section \ref{sec:pretraining}.

\subsection{Results}
\label{sec:result}
\begin{table}[t]
	\caption{
		Performance of our method in the 5-way low-resolution single-domain task.
		``Normalized'' and ``Simple'' mean that the normalized and simple classifiers are used, respectively.
		``All'', ``BN \& FC'', and ``FC'' mean the following:
		the entire network was updated;
		the BN and FC layer were updated;
		only the FC layers were updated.
		``w/o FT'' refers to performance without fine-tuning using novel classes.
		Values with the $\dagger$ mark refer to classification accuracy without fine-tuning,
		as validation accuracy was not increased by fine-tuning the network.
		The * mark means that the classification accuracy for novel classes was not available 
		because the loss value did not decrease in the pretraining phase.
		The - mark means that we did not conduct an experiment
		because the network did not have BN layers.
		\medskip
	}
	\label{table:low_single_domain}
	\centering
	\resizebox{\textwidth}{!}{
		\begin{tabular}{l|cccc|cccc}
			& Normalized All & Normalized BN \& FC & Normalized FC & Normalized w/o FT & Simple All & Simple BN \& FC & Simple FC & Simple w/o FT \\
			\hline
			1-shot ResNet-18 & $48.38\pm0.68^\dagger$ & $48.38\pm0.68^\dagger$ & $48.38\pm0.68^\dagger$ & $48.38\pm0.68$ & $49.27\pm0.68^\dagger$ & $49.27\pm0.68^\dagger$ & $49.27\pm0.68^\dagger$ & $49.27\pm0.68$\\
			1-shot ResNet-34 & $48.41\pm0.69^\dagger$ & $48.41\pm0.69^\dagger$ & $48.41\pm0.69^\dagger$ & $48.41\pm0.69$ & $48.14\pm0.67^\dagger$ & $48.14\pm0.67^\dagger$ & $48.14\pm0.67^\dagger$ & $48.14\pm0.67$\\
			1-shot ResNet-50 & $50.58\pm0.74^\dagger$ & $50.58\pm0.74^\dagger$ & $50.58\pm0.74^\dagger$ & $50.58\pm0.74$ & $49.48\pm0.66^\dagger$ & $49.48\pm0.66^\dagger$ & $49.48\pm0.66^\dagger$ & $49.48\pm0.66$\\
			1-shot ResNet-101 & $48.83\pm0.66^\dagger$ & $48.83\pm0.66^\dagger$ & $48.83\pm0.66^\dagger$ &  $48.83\pm0.66$ & $46.78\pm0.67^\dagger$ & $46.78\pm0.67^\dagger$ & $46.78\pm0.67^\dagger$ & $46.78\pm0.67$\\
			1-shot ResNet-152 & $48.73\pm0.70^\dagger$ & $48.73\pm0.70^\dagger$ & $48.73\pm0.70^\dagger$ & $48.73\pm0.70$ & $49.61\pm0.65^\dagger$ & $49.61\pm0.65^\dagger$ & $49.61\pm0.65^\dagger$ & $49.61\pm0.65$\\
			1-shot VGG-16 & $\bm{54.73\pm0.71}$ & - & $\bm{54.90\pm0.66}$& $49.12\pm0.71$ & * & - & * & *\\
			\hline
			5-shot ResNet-18 & $ 70.81\pm0.54$ & $70.08\pm0.58$ & $64.55\pm0.58^\dagger$  & $64.55\pm0.58$ & $69.50\pm0.58$ & $69.55\pm0.54$ & $68.32\pm0.56$ & $67.19\pm0.56$\\
			5-shot ResNet-34 & $71.90\pm0.55$ & $70.61\pm0.54$ & $65.17\pm0.57^\dagger$ & $65.17\pm0.57$ & $69.48\pm0.61$ & $69.09\pm0.56$ & $67.92\pm0.58$ & $67.23\pm0.60$\\
			5-shot ResNet-50 & $70.80\pm0.57$ & $69.55\pm0.55$ & $66.22\pm0.57^\dagger$ & $66.22\pm0.57$ & $68.92\pm0.58^\dagger$ & $68.92\pm0.58^\dagger$ & $68.92\pm0.58^\dagger$& $68.92\pm0.58$\\
			5-shot ResNet-101 & $69.65\pm0.56$ & $68.81\pm0.56$ & $65.02\pm0.54^\dagger$ & $65.02\pm0.54$ & $65.92\pm0.60^\dagger$ & $65.92\pm0.60^\dagger$ & $65.92\pm0.60^\dagger$& $65.92\pm0.60$\\
			5-shot ResNet-152 & $67.99\pm0.53$ & $66.62\pm0.56$ & $64.48\pm0.54^\dagger$ & $64.48\pm0.54$ & $70.24\pm0.56^\dagger$ & $70.24\pm0.56^\dagger$ & $70.24\pm0.56^\dagger$& $70.24\pm0.56$\\
			5-shot VGG-16 & $\bm{74.50\pm0.50}$ & - & $69.92\pm0.50$ & $66.86\pm0.55$ & * & - & * & *\\
			\hline
		\end{tabular}
	}
\end{table}

\begin{table}[t]
	\caption{
		Performance of our method in the 5-way high-resolution single-domain task.
		``Normalized'' and ``Simple'' mean that the normalized and simple classifiers were used, respectively.
		``All'', ``BN \& FC'', and ``FC'' mean the following:
		the entire network was updated;
		the BN and FC layer were updated;
		only the FC layers were updated.
		``w/o FT'' refers to performance without fine-tuning using novel classes.
		Values with the $\dagger$ mark refer to classification accuracy without fine-tuning,
		as validation accuracy was not increased by fine-tuning the network.
		The * mark means that classification accuracy for novel classes was not available 
		because the loss value did not decrease in the pretraining phase.
		The - mark means that we did not conduct an experiment
		because the network did not have BN layers.
		\medskip
	}
	\label{table:high_single_domain}
	\centering
	\resizebox{\textwidth}{!}{
		\begin{tabular}{l|cccc|cccc}
			& Normalized All & Normalized BN \& FC & Normalized FC & Normalized w/o FT & Simple All & Simple BN \& FC & Simple FC & Simple w/o FT \\
			\hline
			1-shot ResNet-18 & $56.28\pm0.75^\dagger$ & $56.28\pm0.75^\dagger$ & $56.28\pm0.75^\dagger$ & $56.28\pm0.75$ & $57.06\pm0.67^\dagger$ & $57.06\pm0.67^\dagger$ & $57.06\pm0.67^\dagger$ & $57.06\pm0.67$\\
			1-shot ResNet-34 & $56.60\pm0.73^\dagger$ & $56.60\pm0.73^\dagger$ & $56.60\pm0.73^\dagger$ & $56.60\pm0.73$ & $56.83\pm0.70^\dagger$ & $56.83\pm0.70^\dagger$ & $56.83\pm0.70^\dagger$ & $56.83\pm0.70$\\
			1-shot ResNet-50 & $58.05\pm0.73^\dagger$ & $58.05\pm0.73^\dagger$ & $58.05\pm0.73^\dagger$ & $58.05\pm0.73$ & $58.33\pm0.70^\dagger$ & $58.33\pm0.70^\dagger$ & $58.33\pm0.70^\dagger$ & $58.33\pm0.70$\\
			1-shot ResNet-101 & $54.03\pm0.71^\dagger$ & $54.03\pm0.71^\dagger$ & $54.03\pm0.71^\dagger$ & $54.03\pm0.71$ & $51.96\pm0.68^\dagger$ & $51.96\pm0.68^\dagger$ & $51.96\pm0.68^\dagger$ & $51.96\pm0.68$\\
			1-shot ResNet-152 & $56.73\pm0.77^\dagger$ & $56.73\pm0.77^\dagger$ & $56.73\pm0.77^\dagger$ & $56.73\pm0.77$ & $57.61\pm0.71^\dagger$ & $57.61\pm0.71^\dagger$ & $57.61\pm0.71^\dagger$ & $57.61\pm0.71$\\
			1-shot VGG-16 & $59.40\pm0.73$ & - & $\bm{60.88 \pm0.71}$ & $57.15\pm0.72$ & * & - & * & *\\
			\hline
			5-shot ResNet-18 & $78.03\pm0.51$ & $76.87\pm0.52$ & $72.94\pm0.55^\dagger$ & $72.94\pm0.55$ & $76.58\pm0.54$ & $77.20\pm0.49$ & $75.80\pm0.50$ & $75.08\pm0.51$\\
			5-shot ResNet-34 & $77.73\pm0.50$ & $76.85\pm0.50$ & $71.99\pm0.58^\dagger$ & $71.99\pm0.58$ & $77.93\pm0.53$ & $76.15\pm0.47$ & $74.88\pm0.52^\dagger$  & $74.88\pm0.52$\\
			5-shot ResNet-50 & $\bm{79.82\pm0.49}$ & $\bm{79.20\pm0.52}$ & $74.79\pm0.54^\dagger$ & $74.79\pm0.54$ & $77.94\pm0.53^\dagger$ & $77.94\pm0.53^\dagger$ & $77.94\pm0.53^\dagger$ & $77.94\pm0.53$\\
			5-shot ResNet-101 & $78.08\pm0.50$ & $77.82\pm0.48$ & $73.33\pm0.50$ & $71.18\pm0.55$ & $76.62\pm0.52$ & $77.41\pm0.51$ & $78.57\pm0.48$ & $72.32\pm0.53$\\
			5-shot ResNet-152 & $\bm{78.98\pm0.49}$ & $\bm{78.88\pm0.46}$ & $74.57\pm0.55^\dagger$ & $74.57\pm0.55$ & $78.55\pm0.48^\dagger$ & $78.55\pm0.48^\dagger$ & $78.55\pm0.48^\dagger$ & $78.55\pm0.48$\\
			5-shot VGG-16 & $78.80\pm0.50$ & - & $77.48\pm0.49$ & $75.78\pm0.53$ & * & - & * & * \\
			\hline
		\end{tabular}
	}
\end{table}

\begin{table}[t]
	\caption{
		Performance of our method in the 5-way cross-domain task.
		``Normalized'' and ``Simple'' mean that the normalized and simple classifiers were used, respectively.
		``All'', ``BN \& FC'', and ``FC'' mean the following:
		the entire network was updated;
		the BN and FC layer were updated;
		only the FC layers were updated.
		``w/o FT'' refers to performance without fine-tuning using novel classes.
		Values with the $\dagger$ mark refer to classification accuracy without fine-tuning,
		as validation accuracy was not increased by fine-tuning the network.
		The * mark means that classification accuracy for novel classes was not available 
		because the loss value did not decrease in the pretraining phase.
		The - mark means that we did not conduct an experiment
		because the network did not have BN layers.
		\medskip
	}
	\label{table:cross_domain}
	\centering
	\resizebox{\textwidth}{!}{
		\begin{tabular}{l|cccc|cccc}
			& Normalized All & Normalized BN \& FC & Normalized FC & Normalized w/o FT & Simple All & Simple BN \& FC & Simple FC & Simple w/o FT \\
			\hline
			5-shot ResNet-18 & $71.83\pm0.57$ & $68.19\pm0.65$ & $60.26\pm0.62$ & $62.11\pm0.68$ & $68.67\pm0.66$ & $65.59\pm0.62$ & $65.51\pm0.65$ & $64.96\pm0.68$\\
			5-shot ResNet-34 & $71.84\pm0.61$ & $66.93\pm0.64$ & $60.22\pm0.67^\dagger$ & $60.22\pm0.67$ & $71.43\pm0.57$ & $68.36\pm0.59$ & $66.62\pm0.58$ & $67.76\pm0.62$\\
			5-shot ResNet-50 & $\bm{74.88\pm0.58}$ & $70.89\pm0.63$ & $61.45\pm0.68^\dagger$ & $61.45\pm0.68$ & $69.27\pm0.63$ & $67.59\pm0.62$ & $67.77\pm0.66$ & $62.30\pm0.63$\\
			5-shot ResNet-101 & $71.95\pm0.62$ & $68.58\pm0.65$ & $60.94\pm0.57$ & $53.78\pm0.60$ & $65.32\pm0.59$ & $63.93\pm0.61$ & $66.44\pm0.62$ & $49.86\pm0.58$\\
			5-shot ResNet-152 & $73.56\pm0.53$ & $70.16\pm0.62$ & $61.59\pm0.55$ & $61.54\pm0.66$ & $66.90\pm0.64$ & $67.20\pm0.64$ & $68.30\pm0.62$ & $53.45\pm0.63$\\
			5-shot VGG-16 & $71.92\pm0.64$ & - & $65.56\pm0.57$ & $59.84\pm0.64$ & * & - & * & *\\
			\hline
		\end{tabular}
	}
\end{table}
Few-shot classification accuracies for 
the low-resolution single-domain task, high-resolution single-domain task, and cross-domain task
are listed in Tables \ref{table:low_single_domain}, \ref{table:high_single_domain}, and \ref{table:cross_domain},
respectively.

Table \ref{table:low_single_domain} shows that the 
classification accuracy could be increased by approximately $6\%$ 
when the VGG-16 and normalized classifier were used in the 1-shot learning task.
However, the accuracy could not be further improved by updating the entire network in the 1-shot learning task.
However, classification accuracy could be further improved 
by updating the entire network in the 5-shot task.
We assume that this was because the within-class difference could be reduced by fine-tuning the feature extractor
when multiple novel examples were available.

By comparing the results for the high-resolution single-domain (Table \ref{table:high_single_domain})
and low-resolution (Table \ref{table:low_single_domain}) tasks,
it could be argued that the robustness against low-resolution inputs differs depending on the feature extractor.
For example, by comparing the results for 5-shot ``Normalized all'' 
in Table \ref{table:low_single_domain} and \ref{table:high_single_domain},
we can see that the classification accuracy of the ResNet-152 decreased by $11.0\%$
whereas that of the VGG-16 decreased by only $4.3\%$.
This implies that the robustness against low-resolution inputs should also be considered
and that evaluating only few-shot learning performance is difficult.
Although the low-resolution {\it mini}-ImageNet dataset is extremely useful for doing fast experiments,
we must reconsider the validity of the dataset for evaluation of few-shot learning performance.

The comparison of the results from 
the cross-domain task (Table \ref{table:cross_domain}) and high-resolution single-domain task 
shows that the performance decreased in the cross-domain task.
This can be explained by the larger domain shift that occurs between base and novel classes, 
as indicated by \cite{closer_look}.
In addition, the difference in classification accuracy between the cross-domain task and high-resolution single-domain task
was decreased by fine-tuning the entire network.
For example, in the 5-shot learning task using the VGG-16,
the difference in classification accuracy between the single-domain and cross-domain tasks
was $15.9\%$ without fine-tuning,
but it could be reduced to $6.9\%$ by fine-tuning the entire network.
This means that the network could adapt to a large domain shift by having the entire network updated.
We discuss this further in greater detail in Section \ref{sec:increase_performance}.

\subsection{Comparative Evaluation}
\label{sec:comparative_evaluation}
\begin{table}[t]
	\caption{
		Comparison between our method and conventional methods.
		We show several results from different networks in Section \ref{sec:result},
		and therefore in this table show the highest accuracy for each task.
		More specifically, we use the results from
		``VGG-16 Normalized FC'', ``VGG-16 Normalized All'', ``VGG-16 Normalized FC'',
		``ResNet-50 Normalized All'', and ``ResNet-50 Normalized All'' from left to right in this table.
		Values with the $\ddagger$ marks were reported by \cite{closer_look},
		and other values were reported in the original studies.
		The - mark means that the classification accuracy for the task was not reported.
		\medskip
	}
	\label{table:comparison}
	\centering
	\resizebox{\textwidth}{!}{
		\begin{tabular}{l|cc|cc|c}
			& \multicolumn{2}{|c|}{Low-resolution Single-domain} & \multicolumn{2}{|c|}{High-resolution Single-domain} & Cross-domain\\
			& 1-shot & 5-shot & 1-shot & 5-shot & 5-shot\\
			\hline
			{\bf Fine-tune (Ours)} & $54.90\pm0.66$ & $\bm{74.50\pm0.50}$ & $\bm{60.88\pm0.71}$ & $\bm{79.82\pm0.49}$ & $\bm{74.88\pm0.58}$\\
			Baseline \citep{closer_look} & $42.11\pm0.71$ & $62.53\pm0.69$ & $52.37\pm0.79^{\ddagger}$ & $74.69\pm0.64^{\ddagger}$ & $65.57\pm0.70^{\ddagger}$\\
			Baseline++ \citep{closer_look} & $48.24\pm0.75$ & $66.43\pm0.63$ & $53.97\pm0.79^{\ddagger}$ & $75.90\pm0.61^{\ddagger}$ & $62.04\pm0.76^{\ddagger}$\\
			MatchingNet \citep{matching_net}& $46.6$ & $60.0$ & $54.49\pm0.81^{\ddagger}$ & $68.88\pm0.69^{\ddagger}$ & $53.07\pm0.74^{\ddagger}$\\
			ProtoNet \citep{proto_net}& $49.42\pm0.78$ & $68.20\pm0.66$ & $54.16\pm0.82^{\ddagger}$ & $74.65\pm0.64^{\ddagger}$ & $62.02\pm0.70^{\ddagger}$\\
			MAML \citep{MAML}& $48.70\pm1.75$ & $63.15\pm0.91$ & $54.69\pm0.89^{\ddagger}$ & $66.62\pm0.83^{\ddagger}$ & $51.34\pm0.72^{\ddagger}$\\
			RelationNet \citep{relation_net} & $50.44\pm0.82$ & $65.32\pm0.70$ & $53.48\pm0.86^{\ddagger}$ & $70.20\pm0.66^{\ddagger}$ & $57.71\pm0.73^{\ddagger}$\\
			MTL \citep{MTL} & $\bm{61.2\pm1.8}$ & $\bm{75.5\pm0.8}$ & - & - & - \\
			Delta Encoder \citep{delta_encoder} & $\bm{59.9}$ & $69.7$ & - & - & - \\
		\end{tabular}
	}
\end{table}
Table \ref{table:comparison} shows comparative results of ours and previous methods.
Note that we use the best result for each task as given in Section \ref{sec:result} 
because we obtained several results from different networks.
In the 1-shot low-resolution single-domain task,
the classification accuracy was lower than that of the state-of-the-art algorithm,
but it was still higher than those of other common few-shot learning methods such as 
MatchingNet and ProtoNet.
It is interesting to note that our method achieved
nearly the same classification accuracy as that of the state-of-the-art method in the 5-shot task.
The reason for the higher performance in the 5-shot task may be
that the within-class variance could be reduced by fine-tuning the entire network using several examples per class.

In addition, we achieved higher classification accuracy than the reported values of conventional methods
both in the high-resolution single-domain and cross-domain tasks.
The difference in classification accuracy between the 5-shot high-resolution single-domain and cross-domain tasks
was only approximately $5\%$ with our method,
whereas the performance of the conventional methods decreased by more than $10\%$ in the cross-domain task.
This shows that our method successfully reduced the effect of a large domain shift in the cross-domain task
by updating the entire network for novel classes.

\subsection{Increasing Performance by Fine-tuning}
\label{sec:increase_performance}
\begin{figure}[t]
	\begin{minipage}{0.32\hsize}
		\centering
		\includegraphics[width=\linewidth]{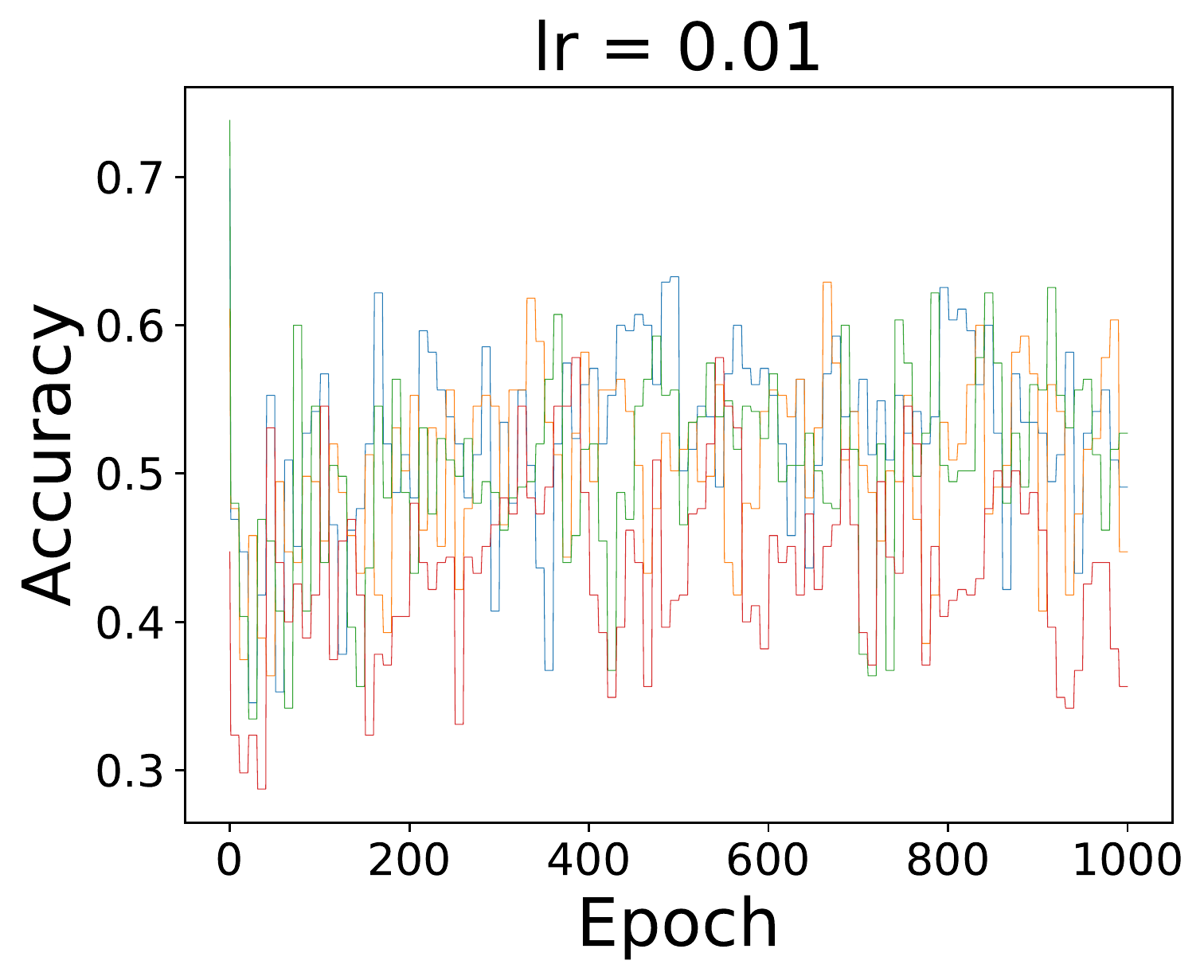}
	\end{minipage}
	\begin{minipage}{0.32\hsize}
		\centering
		\includegraphics[width=\linewidth]{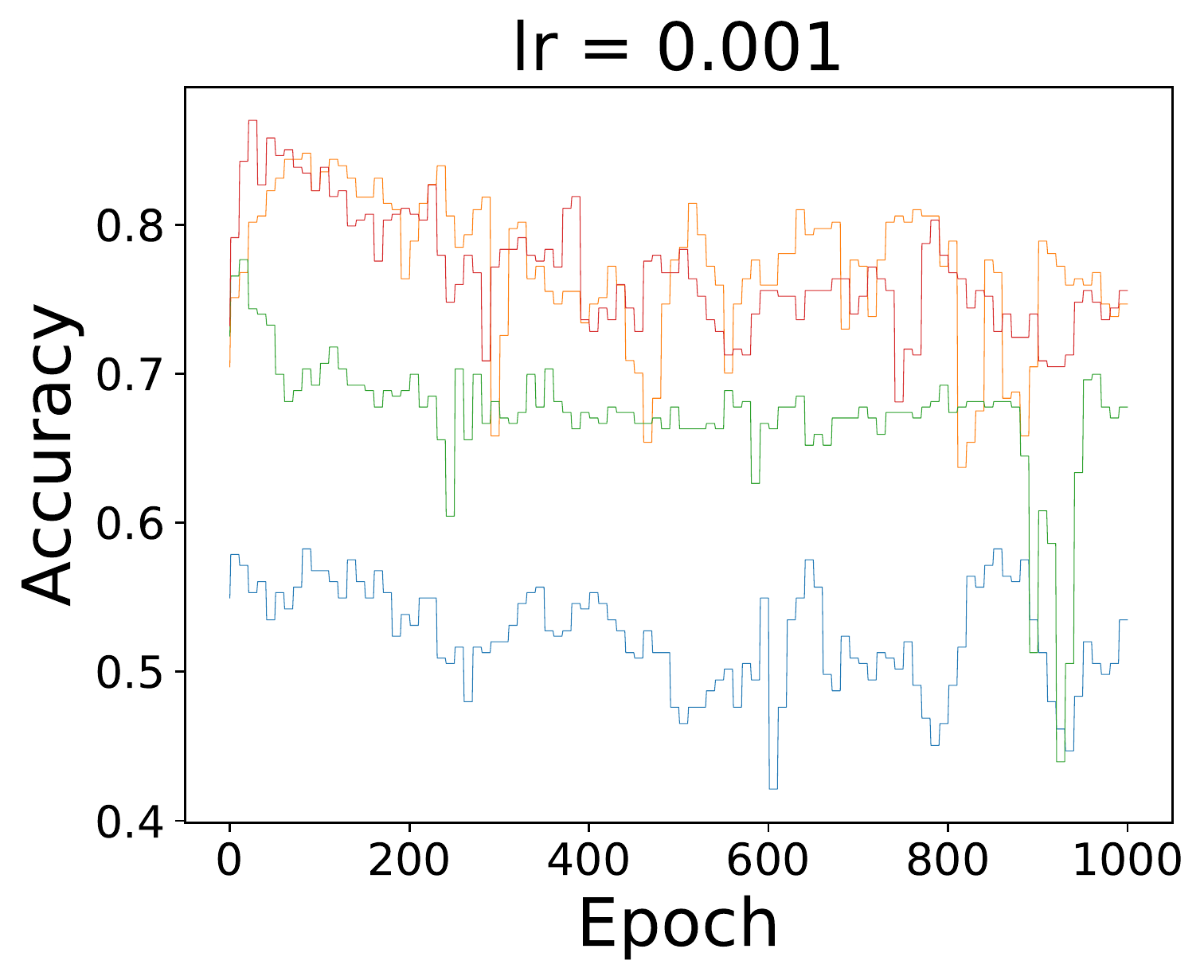}
	\end{minipage}
	\begin{minipage}{0.32\hsize}
		\centering
		\includegraphics[width=\linewidth]{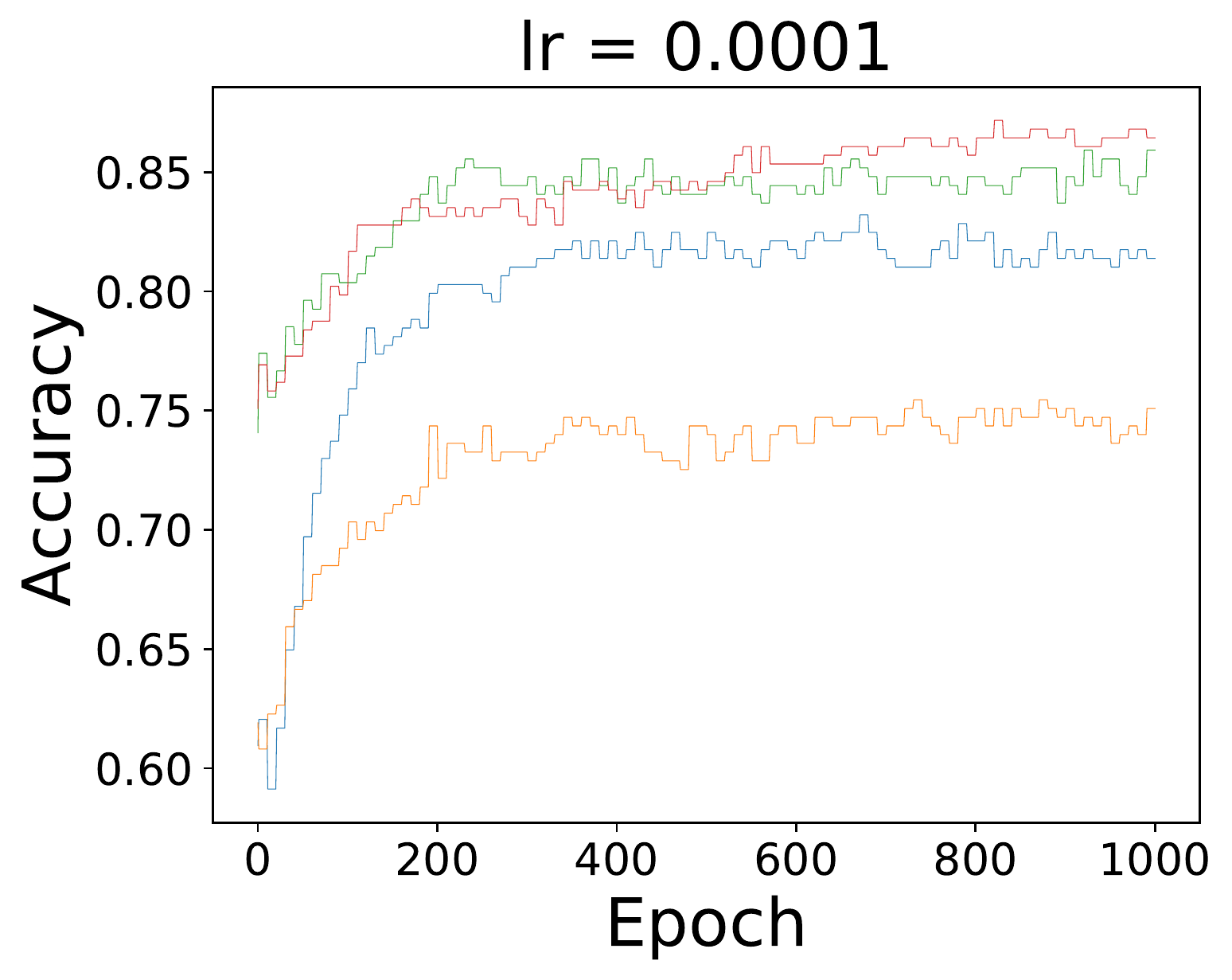}
	\end{minipage}
	\caption{
		Classification accuracy for validation classes with different learning rates.
		We used the ResNet-18, normalized classifier, and Adam optimizer.
		We chose the 5-shot cross-domain task for visualization
		because the transition of validation accuracy is clearer than in other tasks.
		We set the learning rates as $0.01$, $0.001$, and $0.0001$,
		and conducted four trials with randomly selected validation classes and support sets.
		Note that classification accuracy can be significantly changed by the randomly selected classes and samples.
		Therefore, we focused on the transition of the validation accuracy 
		rather than the validation accuracy itself.
	}
	\label{fig:learning_rate}
\end{figure}

\begin{figure}[t]
	\centering
	\includegraphics[width=\linewidth]{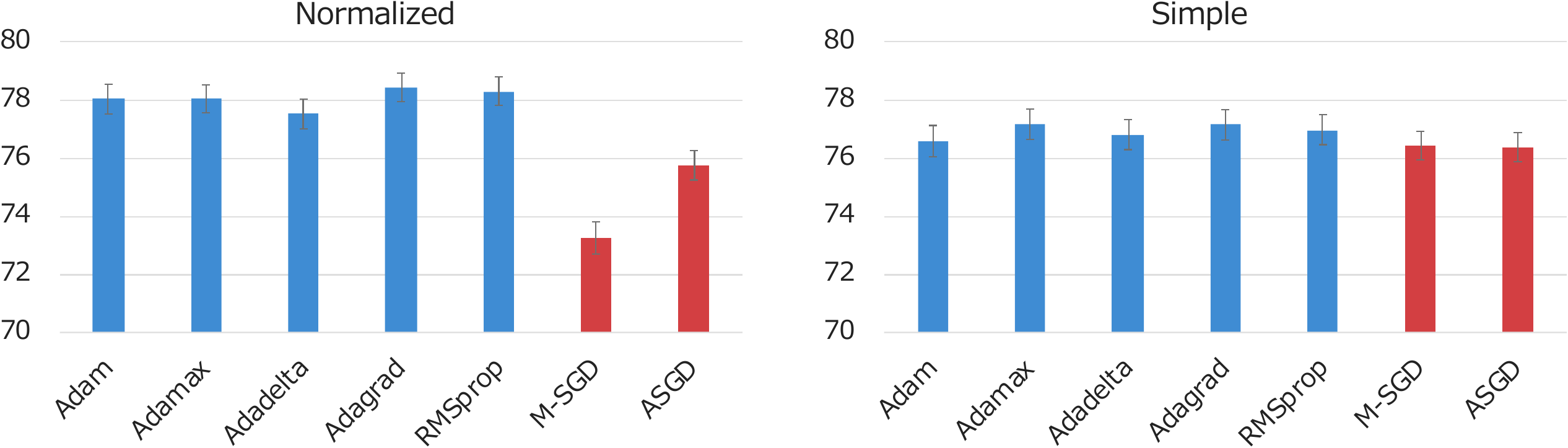}
	\caption{
		Classification accuracy for novel classes with different optimizers in the high-resolution single-domain task.
		We used the ResNet-18 as a feature extractor.
		This shows that the adaptive gradient optimizer (blue bars) achieved higher accuracy
		than other methods (red bars), particularly with the normalized classifier.
	}
	\label{fig:optimizer}
\end{figure}

\begin{figure}[t]
	\begin{minipage}{0.523\hsize}
		\centering
		\includegraphics[height=3.6cm]{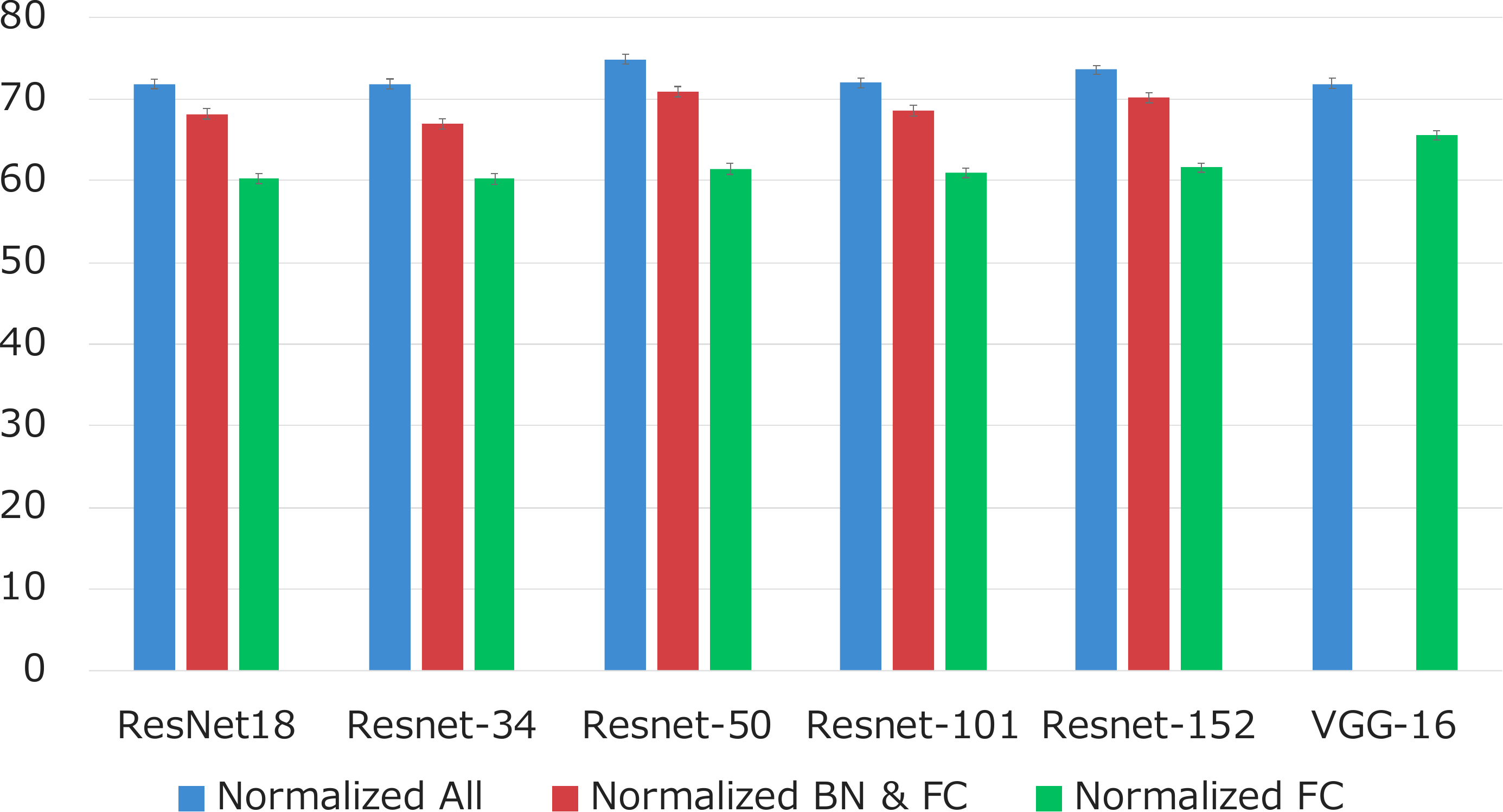}
	\end{minipage}
	\begin{minipage}{0.44\hsize}
		\centering
		\includegraphics[height=3.6cm]{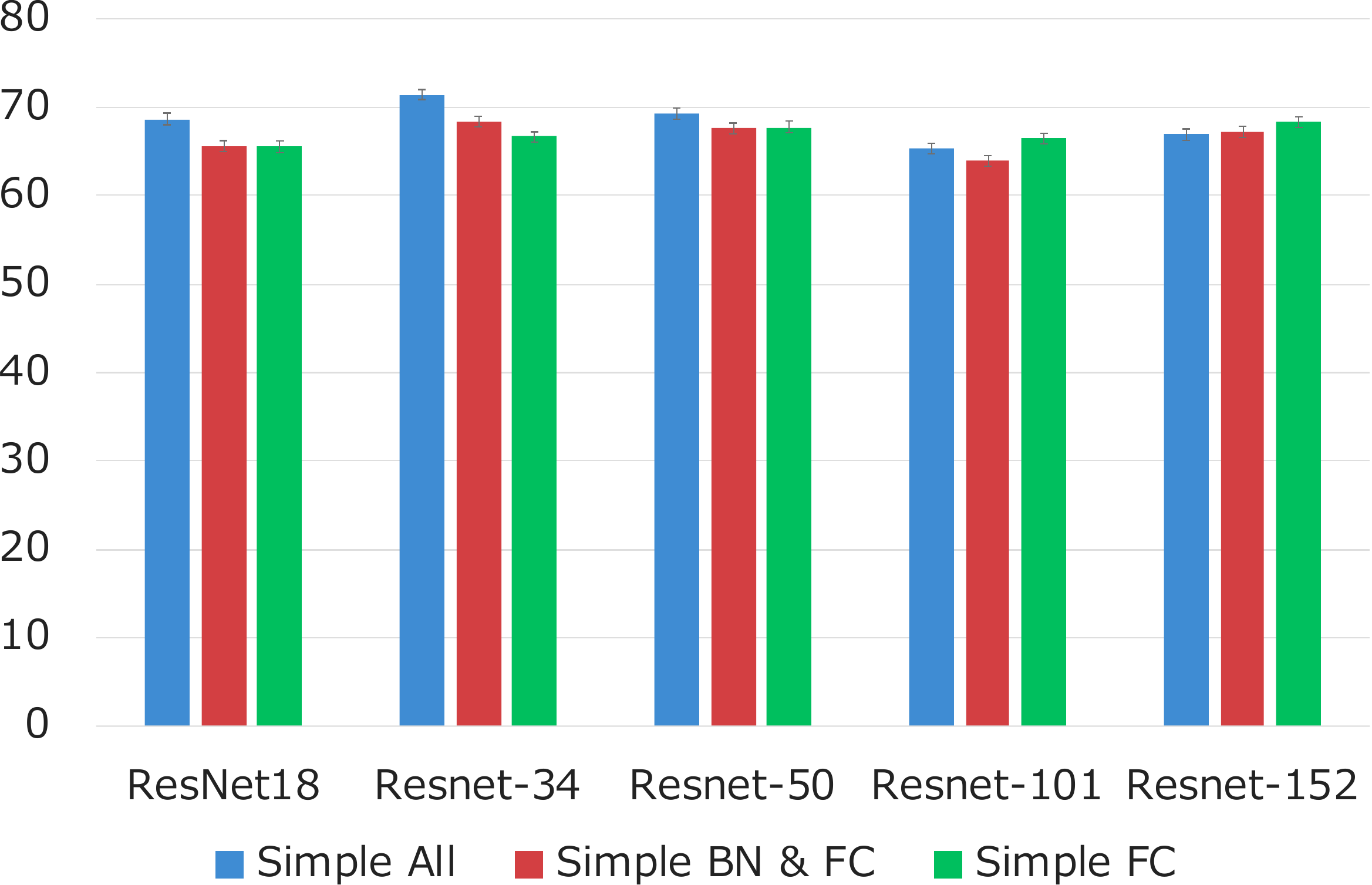}
	\end{minipage}
	\caption{
		Relationship between test accuracy and updated part of the network in the 5-shot cross-domain task.
		We used the Adam optimizer for fine-tuning.
		The result for ``VGG-16 Normalized BN \& FC'' was left empty because the network did not have a BN layer.
		In addition, we did not conduct experiments for the VGG-16 with the simple classifier
		because the loss did not decrease in the pretraining phase.
	}
	\label{fig:cross_domain_comparison}
\end{figure}
We revealed in Section \ref{sec:result} and \ref{sec:comparative_evaluation}
that the fine-tuning method achieved high few-shot classification accuracy in many cases.
In this section, we discuss the means of improving the performance of the fine-tuning method.
We experimentally show that:
\begin{itemize}
	\item Using a low learning rate for fine-tuning stabilizes the retraining process.
	\item Using adaptive gradient optimizers such as Adam increases the classification accuracy.
	\item Higher performance can be obtained by updating the entire network when a large domain shift occurs.
\end{itemize}

{\bf Low Learning Rate}\\
A learning rate is a critical parameter in training a network;
this is also true for fine-tuning for few-shot learning.
Figure \ref{fig:learning_rate} shows that the retraining process can be stabilized
by using a lower learning rate.
For example, the transition of the validation accuracy was unstable when the learning rate was set 
as $0.01$ and $0.001$.
However, the validation accuracy increased in a stable manner when the learning rate was set as $0.0001$,
which is lower than that used in the pretraining phase ($\mathrm{lr}=0.001$).
This means that the learning rate for few-shot fine-tuning should be set low.

{\bf Adaptive Gradient Optimizer}\\
Here, we show that optimizers affect few-shot classification performance
when the network is fine-tuned with only a few examples.
In this experiment, we used the well-known optimizers of 
Adam \citep{Adam-Adamax}, Adamax \citep{Adam-Adamax},
Adadelta \citep{Adadelta}, Adagrad \citep{Adagrad},
RMSprop \citep{RMSprop},
Momentum-SGD, and ASGD \citep{ASGD}.
Of these, Adam, Adamax, Adadelta, Adagrad, and RMSprop are known as adaptive gradient methods.
Figure \ref{fig:optimizer} shows the classification accuracies for a 5-shot high-resolution single-domain task 
using the ResNet-18 with different optimizers.
The results show that higher classification accuracies could be obtained by using adaptive gradient optimizers,
particularly when the normalized classifier was used.
Although \cite{Adam_is_bad} revealed that local minima obtained by the Adam optimizer 
lack a generalization ability,
our experimental results show that this was not necessarily true for few-shot learning.
Why this occurs in few-shot learning is interesting,
but this is beyond the scope of this study.
Therefore, we leave this interesting direction for future works.

{\bf Updating the Entire Network for Adaptation to a Large Domain Shift}\\
Figure \ref{fig:cross_domain_comparison} shows 
the relationship between test accuracy and the updated parts of the network.
This shows that updating the entire network achieves higher accuracy,
particularly when the normalized classifier is used.
Considering the results for the high-resolution single-domain task,
when test accuracy was not further increased by updating the entire network,
it could be argued that updating the entire network when a large domain shift occurs
between base and novel classes is preferable.

\section{Conclusion}
In this study, we showed that in the low-resolution single-domain task,
our fine-tuning method achieved
higher accuracy than common few-shot learning methods in the 1-shot task
and nearly the same accuracy as the state-of-the-art method in the 5-shot task.
We also evaluated our method with more practical tasks, 
such as the high-resolution single-domain and cross-domain tasks.
In both tasks, our method achieved higher accuracy than common few-shot learning methods.
We then experimentally showed that:
1) a low learning rate stabilizes the retraining process,
2) adaptive gradient optimizers such as Adam improve test accuracy, and
3) updating the entire network results in higher accuracy when a large domain shift occurs.
We believe that these insights into fine-tuning for few-shot learning tasks
will help our community tackle this challenging task.

\subsubsection*{Acknowledgments}
This work was partially supported by JST CREST Grant Number JPMJCR1403, 
and partially supported by JSPS KAKENHI Grant Number JP19H01115.
We would like to thank Yusuke Mukuta, Dexuan Zhang, Toshihiko Matsuura, Wataru Kawai,
Hao-Wei Yeh and Atsuhiro Noguchi for their thoughtful comments and suggestions.

\bibliography{iclr2020_conference}
\bibliographystyle{iclr2020_conference}

%\appendix
%\section{Appendix}

\end{document}